\documentclass[letterpaper, 10 pt, conference]{ieeeconf}
\IEEEoverridecommandlockouts
\usepackage{amssymb}
\usepackage{amsmath}
\usepackage{threeparttable}
\usepackage{booktabs}
\usepackage{algorithm}
\usepackage{algorithmicx}
\usepackage{algpseudocode}
\usepackage{xcolor}
\usepackage{graphicx}
\usepackage{subcaption}
\usepackage{multirow}
\usepackage{cite}    
\makeatletter
\let\NAT@parse\undefined
\makeatother
\usepackage{hyperref}
\hypersetup{
hypertex=true,
colorlinks=true,
linkcolor=blue,
anchorcolor=blue,
citecolor=blue
}
\usepackage{xcolor}
\usepackage[table]{xcolor}
\usepackage{pifont}
\usepackage{multirow}
\usepackage{amsfonts}

\begin{document}

\title{\LARGE \bf Cross-Vehicle 3D Geometric Consistency for Self-Supervised \\Surround Depth Estimation on Articulated Vehicles}

\author{Weimin Liu$^{1}$, Jiyuan Qiu$^{2}$, Wenjun Wang$^{1*}$, Joshua H. Meng$^{3}$
\thanks{* Corresponding author: Wenjun Wang.}
\thanks{$^{1}$Weimin Liu is with School of vehicle and Mobility, Tsinghua University, Beijing, China {\tt\small lwm23@mails.tsinghua.edu.cn}}
\thanks{$^{2}$Jiyuan Qiu is with Remote Sensing and Earth Observation Laboratory, University of Copenhagen, Copenhagen K, Denmark {\tt\small jiqi@ign.ku.dk}}
\thanks{$^{1}$Wenjun Wang is with School of vehicle and Mobility, Tsinghua University, Beijing, China {\tt\small wangxiaowenjun@tsinghua.edu.cn}}
\thanks{$^{3}$Joshua H. Meng is with California PATH, University of California, Berkeley, CA, USA {\tt\small hdmeng@berkeley.edu}}
}

\maketitle

\begin{abstract}
Surround depth estimation provides a cost-effective alternative to LiDAR for 3D perception in autonomous driving. While recent self-supervised methods explore multi-camera settings to improve scale awareness and scene coverage, they are primarily designed for passenger vehicles and rarely consider articulated vehicles or robotics platforms. The articulated structure introduces complex cross-segment geometry and motion coupling, making consistent depth reasoning across views more challenging. In this work, we propose \textbf{ArticuSurDepth}, a self-supervised framework for surround-view depth estimation on articulated vehicles that enhances depth learning through cross-view and cross-vehicle geometric consistency guided by structural priors from vision foundation model. Specifically, we introduce multi-view spatial context enrichment strategy and a cross-view surface normal constraint to improve structural coherence across spatial and temporal contexts. We further incorporate camera height regularization with ground plane-awareness to encourage metric depth estimation, together with cross-vehicle pose consistency that bridges motion estimation between articulated segments. To validate our proposed method, an articulated vehicle experiment platform was established with a dataset collected over it. Experiment results demonstrate state-of-the-art (SoTA) performance of depth estimation on our self-collected dataset as well as on DDAD, nuScenes, and KITTI benchmarks.
\end{abstract}

\section{Introduction}
\par Articulated vehicles, such as autonomous rail rapid transit (ART) and other long combination vehicles (LCVs), have become a key component of urban transportation, alleviating congestion and improving transit efficiency in many metropolitans. Despite their operational benefits, the articulated structure of these vehicles introduces significant perception challenges, including enlarged blind spots, an extended perception range, and increased complexity in omnidirectional scene understanding, which are critical for safe navigation and active control. To address these challenges, surround-view perception, such as surround-view depth estimation, which leverages multiple cameras to produce dense 3D scene representations and understanding, has emerged as a promising, low-cost alternative to laser sensors like LiDAR for achieving comprehensive environmental awareness \cite{zhou2017unsupervised}.
\par Depth perception plays a crucial role in 3D scene understanding for autonomous driving. Recent advances in self-supervised monocular approaches allow learning depth directly from raw image sequences without relying on dense groundtruth annotations \cite{yang2021self}\cite{poggi2021synergies}. These methods generally exploit photometric reconstruction within a Structure-from-Motion (SfM) framework. However, these approaches often suffer from scale ambiguity. Extending depth estimation to multi-camera, surround-view setups provides scale-aware full $360^\circ$ coverage, which is critical for holistic scene understanding and multi-view information integration. This extension, however, introduces additional challenges, such as enforcing cross-view consistency and aligning spatial information across cameras. While some recent studies \cite{fsm}\cite{surrounddepth}\cite{cvcdepth} have explored these issues, existing methods rarely capture the full geometric relationships across overlapping views in both spatial and temporal domains. Furthermore, most prior work focuses on standard passenger vehicles, leaving articulated platforms largely unexplored, despite their unique difficulties from dynamic articulation and complex cross-view geometry, which hinder accurate depth reconstruction.
\begin{figure}[t]
    \centering
    \includegraphics[width=1\linewidth]{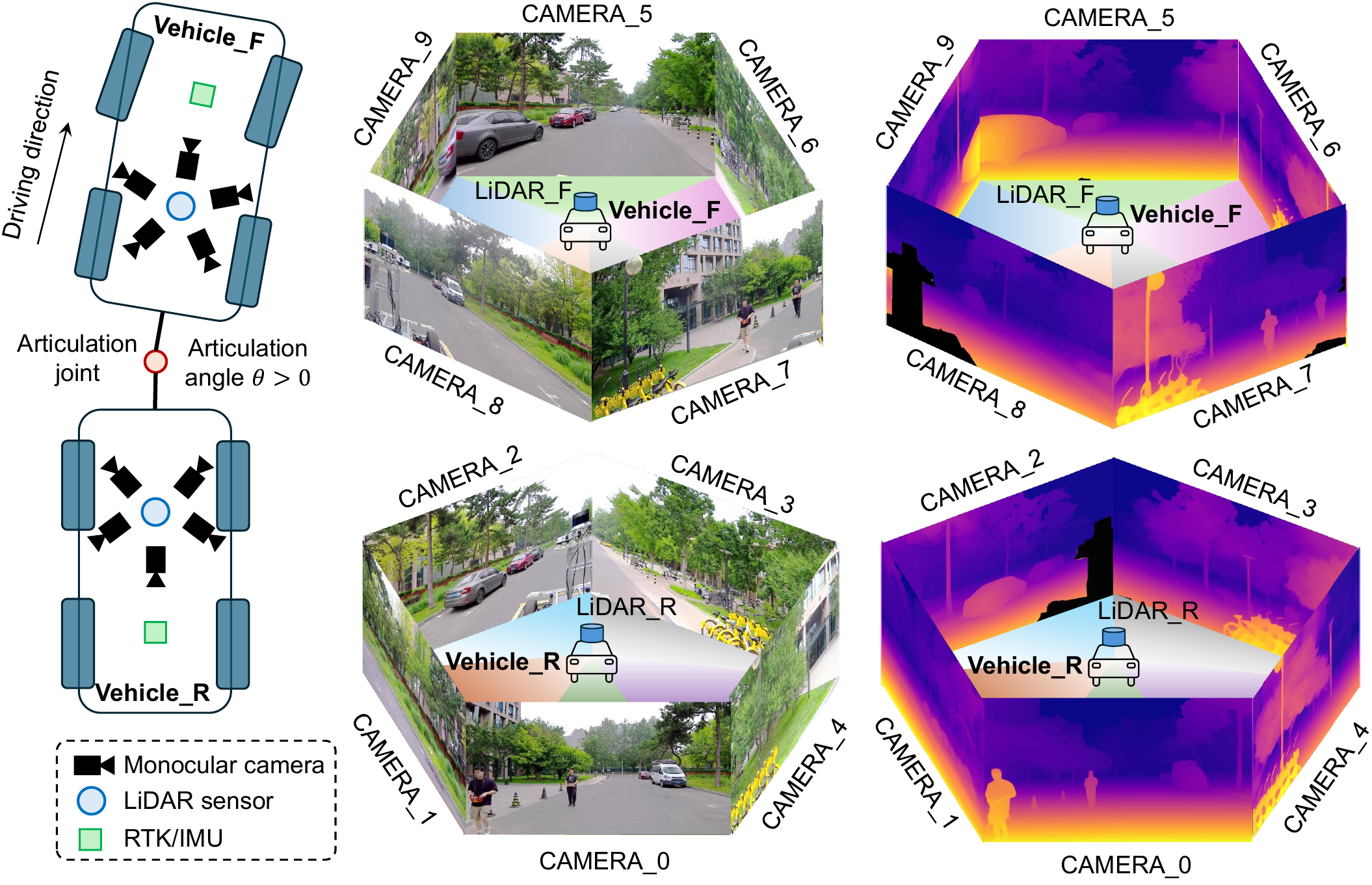}
    \caption{Surround depth estimation for articulated vehicle.}
    \label{fig::abstract}
\end{figure}
\par In this work, we propose \textbf{ArticuSurDepth}, a unified framework and experimental platform for self-supervised surround-view depth estimation on articulated vehicles. Our main contributions are summarized as follows: (1) We establish an experiment platform tailored for articulated vehicles and collect a surround-view dataset to support this task; (2) We introduce a cross-vehicle multi-view spatial context enrichment strategy, which strengthens self-supervision by enhancing both photometric and cross-view geometric consistency; (3) To further enforce 3D structural geometry coherence, we propose a direct interpolation-based cross-view surface normal consistency constraint that mitigates geometric discrepancies across views in spatial and temporal contexts; (4) By leveraging the powerful vision foundation model, we propose a ground plane-aware camera height regularization method that operates in a stable, label-free manner and promotes metric depth estimation; (5) We develop a cross-vehicle pose consistency loss that bridges motion estimation between vehicles, reinforcing geometric coupling across the articulated system.
\par In general, we leverage cross-view, particularly cross-vehicle geometric consistency, as a informative cues to enhance surround-view depth estimation of articulated vehicle by mainly tailoring loss function with incorporation of geometric priors, vehicle articulation, and motion properties.

\section{Related Works}
\par\textbf{Surround-view depth estimation.} Extensive studies have explored surround-view depth estimation for passenger vehicles, aiming to achieve $360^\circ$ dense scene understanding from multi-camera systems. FSM \cite{fsm} first introduces self-supervised depth estimation to the surround-view setting by leveraging photometric reconstruction across spatial and spatial-temporal contexts together with multi-camera pose consistency. Building upon this paradigm, subsequent works mainly focus on enhancing cross-view interaction and motion modeling within multi-camera systems. For example, VFDepth \cite{vfdepth} adopts unified volumetric feature fusion and canonical motion estimation to impose global motion constraints, while SurroundDepth \cite{surrounddepth} strengthens multi-view feature interaction through joint motion estimation and a Cross-View Transformer. More recently, MCDP \cite{mcdp} incorporates pseudo-depth generated by the pretrained DepthAnything V1 model \cite{depthanythingv1} and performs conditional denoising refinement to further improve depth estimation. In addition, CVCDepth \cite{cvcdepth} and GeoSurDepth \cite{liugeosurdepth} further enhance performance by enforcing geometric consistency across views.
\par\textbf{Surround-view perception for articulated vehicles.}
Compared with perception systems for passenger vehicles, research on articulated vehicles remains relatively underexplored due to their complex kinematic structures and time-varying cross-vehicle geometry. ArticuBEVSeg \cite{liu2025articubevseg} proposes an end-to-end articulation-aware BEV road semantic understanding framework for articulated LCVs, which integrates distorted surround-view fisheye images with time-varying cross-vehicle extrinsics. Feng \textit{et al.} \cite{feng2019calibration} extend panoramic view generation from passenger vehicles to articulated platforms by rotating BEV images according to articulation angles and fusing them across vehicles. To mitigate photometric degradation and information loss in the transition regions of adjacent side-mounted cameras during turning maneuvers, Liu \textit{et al.} \cite{liu2026weak} propose a weak-supervised framework for joint panoramic view generation and articulation angle estimation for a three-carbody articulated LCV, enabling articulation-angle-independent holistic surround-view perception while reducing blind spots.

\section{Problem Formulation and Overall Framework}
\begin{figure*}
    \centering
    \includegraphics[width=1\linewidth]{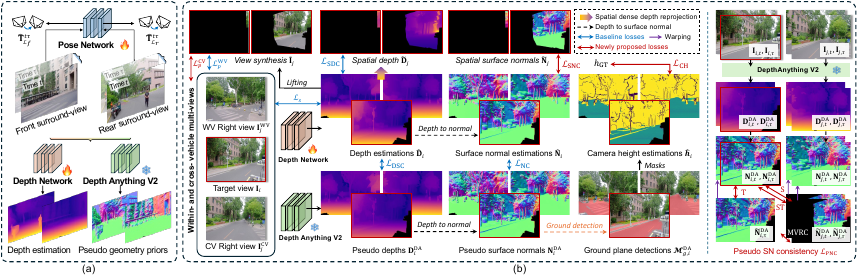}
    \caption{\textbf{Overview:} (a) Network architecture of ArticuSurDepth; (b) Self-supervised training framework and its loss components: (Left) Within- and cross-vehicle spatial context enrichment. Example: for the target view $C_5$, the within-vehicle right view is $C_6$, while the type-2 cross-vehicle right view is $C_0$. (Right) Cross-view pseudo surface normal consistency ($\mathcal{L}_\text{PNC}$).}
    \label{fig::overview}
\end{figure*}
\par We formulate self-supervised surround-view depth estimation for articulated vehicles under the SfM paradigm, jointly learning dense depth and ego-motion from multi-camera sequences. As shown in Fig. \ref{fig::overview}(a), our framework integrates trainable depth and pose networks with the frozen vision foundation model DepthAnything (DA) V2 \cite{depthanythingv2}, which provides pseudo geometric priors to facilitate 3D structure learning and improve depth estimation.
\par During training, the depth network predicts depth maps $\{\hat{\mathbf{D}}_i\}_{i=1}^N$ from surround-view images $\{\mathbf{I}_i^t\}_{i=1}^N$ at target time $t$. DA provides pseudo geometric priors in temporal and spatial contexts, which serve as guidance for subsequent 3D reconstruction, view synthesis, and geometric consistency learning. Meanwhile, the pose network takes consecutive surround-view frames $\{(\mathbf{I}_i^t,\mathbf{I}_i^{\tau})\}_{i=1}^N$ to estimate joint relative motion, which is subsequently distributed to each camera via offline calibrated extrinsics, yielding $\{\hat{\mathbf{T}}_i^{t\tau}\in \text{SE}(3)\}_{i=1}^N$, where $\tau$ indicates source time. The estimated depth and pose are jointly used for spatial and temporal warping, providing photometric and geometric supervisions. 
\par To verify the effectiveness of our proposed framework, an experiment platform was established where a dataset was collected (see Fig. \ref{fig::abstract} and Fig. \ref{fig::platform}). The overall training is performed in a fully self-supervised fashion, without leveraging groundtruth or pseudo depth as direct supervision signals. Notably, both the pose network and DA are utilized during training and discarded at inference. The proposed modules, loss functions, experiment setup and results are detailed elaborated in subsequent sections.

\section{Method}
\label{sec::method}
\subsection{Cross-Vehicle Multi-View Spatial Context Enrichment}
\label{sec::context}
\par
\begin{figure}[h]
    \centering
    \includegraphics[width=1\linewidth]{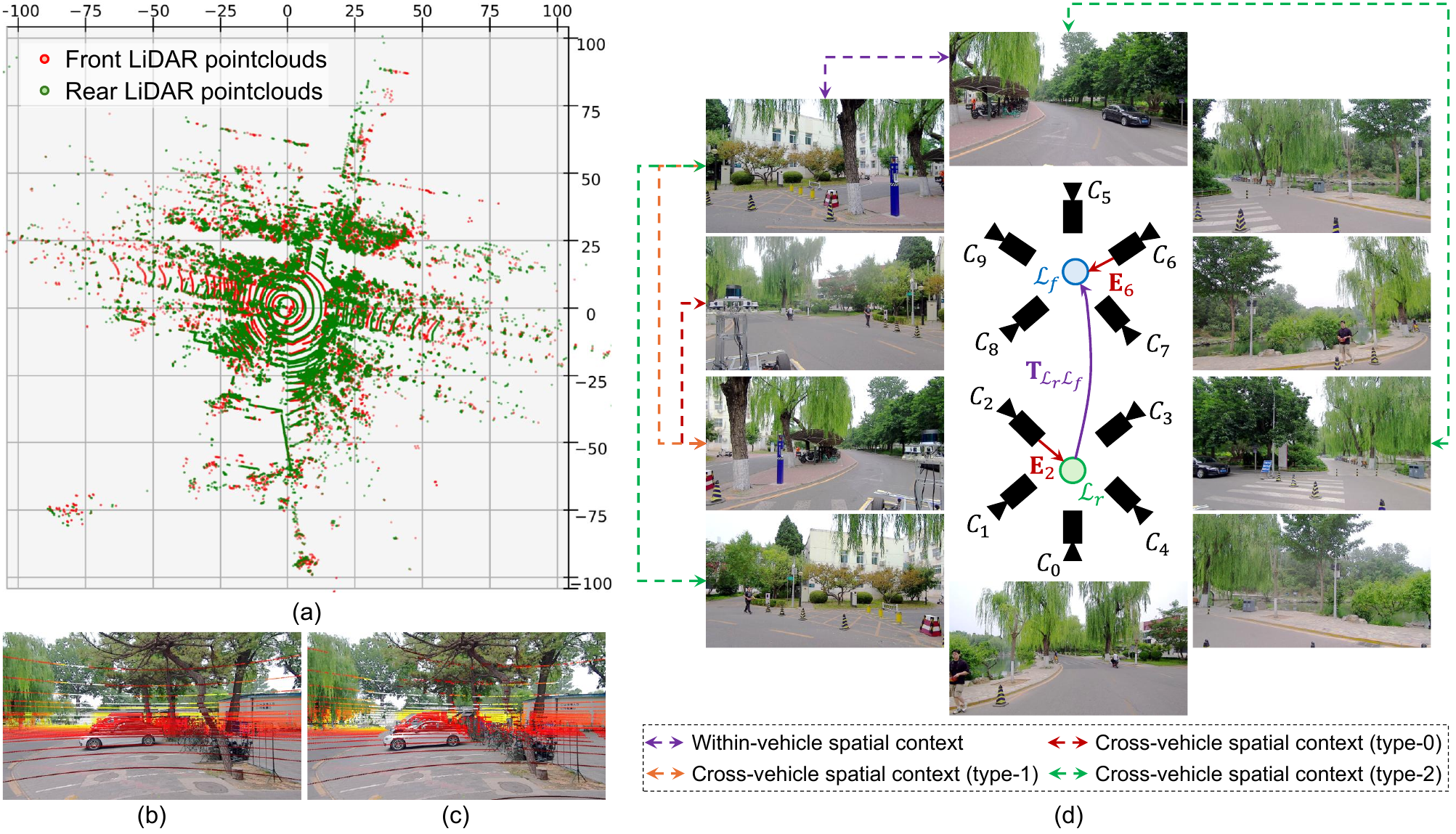}
    \caption{Example of cross-vehicle extrinsics calibration: (a) LiDAR pointclouds registration; (b) Pointclouds of $\mathcal{L}_f$ projected on camera $C_6$ (mounted on front vehicle); (c) Pointclouds of $\mathcal{L}_r$ projected on camera $C_6$; (d) Within- and cross-vehicle spatial contexts and transformations.}
    \label{fig::calibration}
\end{figure}

 Photometric loss forms the foundation of self-supervised depth learning by minimizing the discrepancy between the target image and its synthesized views. While conventional approaches primarily exploit temporal correspondence and further incorporate spatial and spatial-temporal contexts \cite{fsm} to strengthen geometric constraints and improve metric depth estimation, we extend these contextual relationships across vehicles, introducing additional overlapping views and richer supervisory signals for LCVs. The image reconstruction via pixel-wise warping is formulated as,
\begin{equation}
    \mathbf{p}_{ij}^{t\tau}=\mathbf{\Pi}_{ij}^{t\tau}\mathbf{p}_i^t,~\tilde{\mathbf{I}}_{ij}^{t\tau}(\mathbf{p})=\left<\mathbf{I}_j^{\tau}\right>_{\mathbf{p}_{ij}^{t\tau}},~\mathbf{\Pi}_{ij}^{t\tau}=\mathbf{K}_j\mathbf{X}_{ij}^{t\tau}\hat{\mathbf{D}}_i\mathbf{K}_i^{-1},
    \label{eq::project}
\end{equation}
\begin{equation}
    \mathbf{X}_{ij}^{t\tau}= \begin{cases}
    \hat{\mathbf{T}}_i^{t\tau}, & \text{temporal context}, \\ 
    \mathbf{E}_j \mathbf{E}_i^{-1}, & \text{WV spatial context}, \\ 
    \mathbf{E}_j\mathbf{T}_{\mathcal{L}_i\mathcal{L}_j}\mathbf{E}_i^{-1}, & \text{CV spatial context}, \\ 
    \hat{\mathbf{T}}_j^{t \tau}\mathbf{E}_j \mathbf{E}_i^{-1}, & \text{WV spatial-temporal context},\\
    \hat{\mathbf{T}}_j^{t \tau}\mathbf{E}_j\mathbf{T}_{\mathcal{L}_i\mathcal{L}_j}\mathbf{E}_i^{-1}, & \text{CV spatial-temporal context},
    \end{cases}
    \label{eq::context}
\end{equation}
where WV and CV indicate within-vehicle and cross-vehicle, respectively. $\left<\right>$ implies bilinear sampling. $\mathbf{E}$ and $\mathbf{K}$ indicate extrinsics and intrinsics matrices calibrated offline. Notably, all extrinsics here refer to the transformation from corresponded camera to LiDAR coordinate system. $\mathbf{T}_{\mathcal{L}_i\mathcal{L}_j}$ denotes the cross-vehicle extrinsics transformation from the LiDAR coordinate frame of camera $i$ to that of camera $j$. Instead of relying on rough and coarse manual physical measurements for calibration, we estimate the transformation by registering the two LiDAR pointclouds using the Iterative Closest Point (ICP) algorithm. In Fig. \ref{fig::calibration}(a)-(c), we present visualization of a calibration example, where it shows accurately aligned pointcloud registration and cross-vehicle pointcloud-to-camera projection.
\par The reconstruction error is computed as a weighted sum of intensity differences and structural similarity \cite{godard2017unsupervised}\cite{godard2019digging},
\begin{equation}
    pe(\mathbf{x},\mathbf{y})=(1-\alpha)\Vert\mathbf{x}-\mathbf{y}\Vert_1+\alpha\frac{1-\text{SSIM}(\mathbf{x},\mathbf{y})}{2},
\end{equation}
where $\alpha$ is the weighting coefficient, and $\alpha=0.85$.
\par For each context used in pixel warping, the corresponding photometric loss is defined as follows, and the overall loss is obtained by aggregating across all contexts.
\begin{equation}
    \begin{cases}
    \mathcal{L}_p^\text{T}=\min_{\tau}pe(\mathbf{I}_i^t,\tilde{\mathbf{I}}_i^{\tau}),&\text{temporal context}, \\ 
    \mathcal{L}_p^\text{S}=pe(\mathbf{I}_i^t,\tilde{\mathbf{I}}_j^{t}),&\text{spatial context}, \\ \mathcal{L}_p^\text{ST}=\min_{\tau}pe(\mathbf{I}_i^t,\tilde{\mathbf{I}}_j^{\tau}),&\text{spatial-temporal context},\\
    \mathcal{L}_p^\text{MVRC}=\min_{\tau}pe(\tilde{\mathbf{I}}_j^t,\tilde{\mathbf{I}}_j^{\tau}),&\text{MVRC},\\
    
    \end{cases}
\end{equation}
\begin{equation}
    \mathcal{L}_p=\lambda_\text{T}\mathcal{L}_p^\text{T}+\lambda_\text{S}\mathcal{L}_p^\text{S}+\lambda_\text{ST}\mathcal{L}_p^\text{ST}+\lambda_\text{MVRC}\mathcal{L}_p^\text{MVRC},
\end{equation}
where $\lambda\_$ indicates weighting coefficient. MVRC refers to the multi-view reconstruction consistency loss proposed by CVCDepth \cite{cvcdepth}, which calculates the photometric error of reconstructed images generated with spatial and spatial-temporal contexts.
\par
\begin{figure}[t]
    \centering
    \includegraphics[width=1\linewidth]{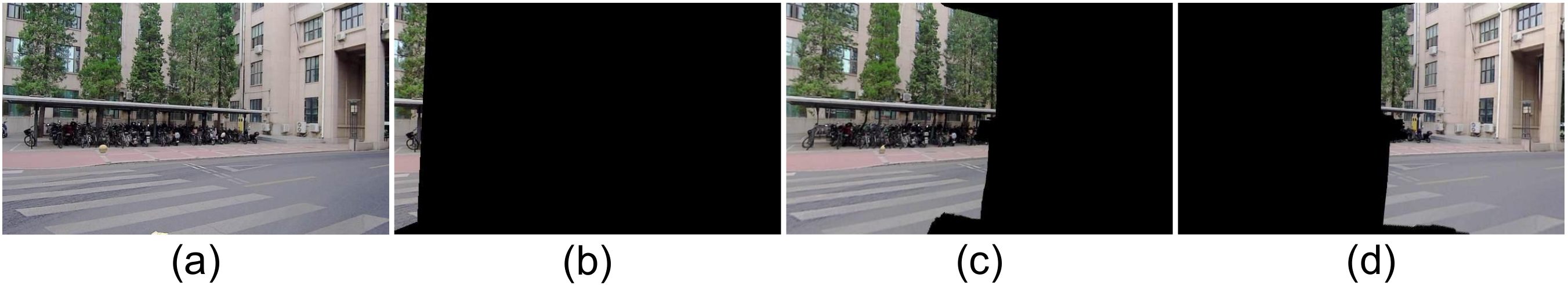}
    \caption{Example of spatial warps: (a) Color image of $C_5$; (b) Spatial warp from left camera $C_9$; (c) Spatial cross-vehicle warp (type-2) from left camera $C_2$; (d) Spatial cross-vehicle warp (type-2) from right camera $C_3$.}
    \label{fig::context}
\end{figure}

 Fig. \ref{fig::calibration}(d) presents examples of within-vehicle and cross-vehicle spatial contexts and transformations. The introduced CV spatial contexts are expected to compensate for the rather limited overlapping FoV inherent to within-vehicle cross-view spatial contexts. Unlike WV contexts, which assume fixed cross-view transformations, CV contexts are dynamically determined by the relative poses between vehicles. Specifically, we categorize CV contexts into three types based on the number of intermediate views between paired cameras. In practice, type-1 and type-2 CV contexts generally provide broader overlapping FoVs, whereas type-0 contexts may exhibit increased overlap under specific motions such as turning. The proposed cross-vehicle spatial context enrichment is applied not only to view synthesis and image reconstruction, but also to all spatial warping operations involving cross-vehicle transformations. In Fig. \ref{fig::context} we present an example. Table \ref{table::cv_context} lists out camera pairs for all cross-vehicle spatial contexts used in this study. 

\begin{table}[t]
\renewcommand{\arraystretch}{1}
\setlength{\tabcolsep}{2pt}
\centering
\caption{List of camera pairs for cross-vehicle spatial contexts.}
\label{table::cv_context}
\scalebox{0.93}{
\begin{tabular}{c|c}
\toprule
Type & Camera pairs\\
\midrule
Type-0 & $(C_8\&C_2)$, $(C_7\&C_3)$\\
Type-1 & $(C_9\&C_2)$, $(C_6\&C_3)$, $(C_8\&C_1)$, $(C_7\&C_4)$\\
Type-2 & $(C_5\&C_2)$, $(C_5\&C_3)$, $(C_9\&C_1)$, $(C_6\&C_4)$, $(C_8\&C_0)$, $(C_7\&C_0)$\\
\bottomrule
\end{tabular}
}
\end{table}

\subsection{Cross-View Surface Normal Consistency}
\subsubsection{Cross-view geometry consistency} Cross-view geometry consistency constitutes an important role in enhancing surround-view depth estimation performance. CVCDepth \cite{cvcdepth} proposed a spatial depth consistency loss by calculating the difference between the depth estimation in target view and that reprojected from spatial adjacent views through a modified backward warping strategy,
\begin{equation}
    \mathbf{P}_{ji}=\mathbf{E}_i\mathbf{E}_j^{-1}\hat{\mathbf{D}}_j(\mathbf{p}_j)\mathbf{K}_j^{-1}\mathbf{p}_j,~\tilde{\mathbf{D}}_j(\mathbf{p})=\left<(\mathbf{P}_{ji})_z\right>_{\mathbf{p}_{ij}},
    \label{eq::lift}
\end{equation}
where $(\mathbf{P}_{ji})_z$ denotes $z$ value of a 3D point reprojected from coordinate system of camera $j$ to that of camera $i$. $\tilde{\mathbf{D}}_j$ implies reprojected spatial depth from adjacent view $j$. The spatial depth consistency loss can thus be formulated as follows,
\begin{equation}
    \mathcal{L}_\text{SDC}=\Vert\hat{\mathbf{D}}_i-\tilde{\mathbf{D}}_j\Vert_1.
\end{equation}
\par In addition to depth consistency, 3D structural geometry consistency further facilitates depth estimation in an implicit manner, where surface normal provides a compact representation of local geometric structure. In \cite{liugeosurdepth}, the authors demonstrate that surface normal derived from either depth or pseudo-depth are invariant to scale and drift. Based on this property, they formulate surface normal consistency (NC) using the inner product between normalized surface normals as follows, 
\begin{equation}
    NC(\mathbf{N}_x,\mathbf{N}_y)=1-\Vert\mathbf{N}_x^\top\mathbf{N}_y\Vert_1.
\end{equation}
\par Building upon this formulation, \cite{liugeosurdepth} extends the scale- and drift-invariance property of surface normal consistency from the target view to the spatial context and proposes a \textbf{depth interpolation-based} surface normal reprojection strategy. In this approach, the surface normal in the target view reprojected from spatial context is computed through the following steps: (1) reprojecting depth or pseudo-depth from the source context; (2) lifting pixels into 3D space; (3) applying pose transformation; (4) reconstructing a depth map in the target coordinate system via bilinear sampling; and (5) computing surface normals from the reconstructed depth map. The NC losses in the target view and spatial context are formulated as,
\begin{equation}
    \mathcal{L}_\text{NC}=NC(\hat{\mathbf{N}}_i,\mathbf{N}_i^\text{DA}),~\tilde{\mathcal{L}}_\text{NC}=NC(\tilde{\mathbf{N}}_j,\tilde{\mathbf{N}}_j^\text{DA}),
\end{equation}
where $\hat{\mathbf{N}}_i$ and $\mathbf{N}_i^\text{DA}$ indicates surface normals generated with estimated depth and pseudo depth of DA in target view $i$. $\tilde{\mathbf{N}}_j$ and $\tilde{\mathbf{N}}_j^\text{DA}$ implies surface normals generated with reprojected spatial estimated depth and pseudo depth of DA from adjacent view $j$.
\par Motivated by aforementioned studies, in this work, we move beyond the computation method in spatial context and instead propose a \textbf{direct interpolation-based} method, while preserving the geometric closure of the reprojection constraint. Specifically, in this work, we obtain surface normal from source context directly and further constitute cross-view surface normal consistency. 
\subsubsection{Direct interpolation-based surface normal reprojection.} The surface normal at a point on a surface is defined as the normalized cross product of two tangent vectors in 3D space. However, since images are inherently discrete, the continuous tangent vectors cannot be directly obtained. Therefore, in studies like \cite{liugeosurdepth}\cite{xue2020toward}, the tangent directions are approximated using 3D offset vectors between neighboring pixels satisfying $\cos(\overrightarrow{\mathbf{pp}^x},\overrightarrow{\mathbf{pp}^y})=0$. The surface normal at pixel $\mathbf{p}$ can be written as,
\begin{equation}
    \mathbf{N}(\mathbf{p}):=\frac{\mathbf{v}^x\times \mathbf{v}^y}{\Vert \mathbf{v}^x\times \mathbf{v}^y \Vert}
    \approx\frac{\overrightarrow{\mathbf{P}\mathbf{P}^x}\times\overrightarrow{\mathbf{P}\mathbf{P}^y}}{\Vert\overrightarrow{\mathbf{P}\mathbf{P}^x}\times\overrightarrow{\mathbf{P}\mathbf{P}^y}\Vert},
    \label{eq::norm}
\end{equation}
where $\mathbf{v}^x$ and $\mathbf{v}^y$ indicate tangent vectors. $\mathbf{v}^x=\partial\mathbf{P}/{\partial x}$ and $\mathbf{v}^y=\partial\mathbf{P}/{\partial y}$. 3D projection can be obtained via $\mathbf{P}=D\mathbf{K}^{-1}\mathbf{p}$.
\par To illustrate how our proposed surface normal reprojection work across different contexts, here we use spatial context as an example. Given target view $i$ and adjacent view $j$, the 3D point reconstructed from view $i$ can be transformed to view $j$ via spatial transformation $\mathbf{T}_{ij}$, where $\mathbf{T}_{ij}=[\mathbf{R}_{ij},\mathbf{t}_{ij}]=\mathbf{E}_j\mathbf{E}_i^{-1}$. Under this transformation, the tangent vectors in view $j$ can be formulated as follows,
\begin{equation}
    \begin{aligned}
        &\mathbf{v}_{ij}^x=\frac{\partial \mathbf{P}_{ij}}{\partial x}=\frac{\partial(\mathbf{R}_{ij}\mathbf{P}_i+\mathbf{t}_{ij})}{\partial x}=\mathbf{R}_{ij}\frac{\partial\mathbf{P}_i}{\partial x}=\mathbf{R}_{ij}\mathbf{v}_i^x,\\
        &\mathbf{v}_{ij}^y=\mathbf{R}_{ij}\mathbf{v}_i^y.
    \end{aligned}
    \label{eq::tangent}
\end{equation}

\par Since $(\mathbf{Ra})\times(\mathbf{Rb})=\mathbf{R}(\mathbf{a}\times\mathbf{b})$ and $\mathbf{R}^\top\mathbf{R}=\mathbf{I}$, by substituting (\ref{eq::tangent}) into (\ref{eq::norm}), the surface normal in view $j$ can be written as,
\begin{equation}
    \begin{aligned}
        \mathbf{N}_j(\mathbf{p}_{ij})&=\frac{\mathbf{v}_{ij}^x \times \mathbf{v}_{ij}^y}{\Vert \mathbf{v}_{ij}^x \times \mathbf{v}_{ij}^y \Vert}\\
        &=\frac{(\mathbf{R}_{ij}\mathbf{v}_i^x)\times (\mathbf{R}_{ij}\mathbf{v}_i^y)}{\Vert (\mathbf{R}_{ij}\mathbf{v}_i^x)\times(\mathbf{R}_{ij}\mathbf{v}_i^y)\Vert}=\mathbf{R}_{ij}\hat{\mathbf{N}}_i(\mathbf{p}_i),
    \end{aligned}
\end{equation}
where $\mathbf{p}_{ij}=\mathbf{K}_j\mathbf{T}_{ij}\hat{\mathbf{D}}_i(\mathbf{p}_i)\mathbf{K}_i^{-1}\mathbf{p}_i$, which indicates that surface normal in adjacent view is theoretically consistent with the rotated surface normal from target view. This property reflects that surface normals are invariant to translation. 
\par In practice, the reprojected pixel $\mathbf{p}_{ij}$ generally falls at a sub-pixel location. Therefore, the corresponding surface normal in view $j$ is obtained via bilinear sampling on the estimated surface normal map, i.e., $\tilde{\mathbf{N}}_j(\mathbf{p}_{ij}) := \left< \hat{\mathbf{N}}_j\right>_{\mathbf{p}_{ij}}$. Based on this, we define the cross-view surface normal consistency in the spatial context (SNC) as,
\begin{equation}
    \mathcal{L}_\text{SNC}=NC(\mathbf{R}_{ij}^\top\tilde{\mathbf{N}}_j,\hat{\mathbf{N}}_i),
\end{equation}
which formulates consistency between surface normal in target view and interpolated surface normal from adjacent view but compensated with the rotation in perspective transformation.

\subsubsection{Cross-view pseudo surface normal consistency.} Given that surround-view images contain multiple views captured over time, we further generalize NC to more structural priors on various contexts. Specifically, by following aforementioned direct interpolation-based surface normal reprojection strategy and formulation pattern of photometric losses, we further propose a cross-view pseudo surface normal consistency (PNC) to provide supervision signal for surround-view joint depth and motion estimation. For notation simplicity, we denote $\check{\mathbf{N}}$ as the interpolated surface normal with rotation compensation, which can be formulated in temporal, spatial, and spatial-temporal contexts as follows,
\begin{equation}
    \begin{aligned}
        & \check{\mathbf{N}}_{i,\tau}^\text{T}=\hat{\mathbf{R}}_{i,{t\tau}}^\top\tilde{\mathbf{N}}_{i,\tau}^\text{DA},\\
        & \check{\mathbf{N}}_{j,t}^\text{S}=\mathbf{R}_{ij,t}^\top\tilde{\mathbf{N}}_{j,t}^\text{DA},\\
        & \check{\mathbf{N}}_{j,\tau}^\text{ST}=(\hat{\mathbf{R}}_{j,{t\tau}}\mathbf{R}_{ij,t})^\top\tilde{\mathbf{N}}_{j,\tau}^\text{DA}.\\
    \end{aligned}
\end{equation}
\par The PNC in various contexts can thus be formulated as follows,
\begin{equation}
    \begin{cases}
    \mathcal{L}_\text{PNC}^\text{T}=\min_\tau NC(\check{\mathbf{N}}_{i,\tau}^\text{T},\mathbf{N}_{i,t}^\text{DA}),&\text{temporal context}, \\ 
    \mathcal{L}_\text{PNC}^\text{S}=NC(\check{\mathbf{N}}_{j,t}^\text{S},\mathbf{N}_{i,t}^\text{DA}),&\text{spatial context}, \\ \mathcal{L}_\text{PNC}^\text{ST}=\min_\tau NC(\check{\mathbf{N}}_{j,\tau}^\text{ST},\mathbf{N}_{i,t}^\text{DA}),&\text{spatial-temporal context},\\
    \mathcal{L}_\text{PNC}^\text{MVRC}=\min_{\tau}NC(\check{\mathbf{N}}_{j,t}^\text{S}, \check{\mathbf{N}}_{j,\tau}^\text{ST}),&\text{MVRC}.\\
    
    \end{cases}
\end{equation}

\subsubsection{Comparison between depth and direct interpolation-based surface normal reprojection methods.} Both approaches rely on pixel correspondence and bilinear sampling. However, they differ in the order of interpolation. From a geometric perspective, two methods are not equivalent since surface normal estimation is a nonlinear operation involving cross products of tangent offset vectors. 
\par In our proposed surface normal reprojection method, surface normal from source context reprojected in target view are computed at discrete pixels and then interpolated. Here we let the four neighboring pixels of a reprojected continuous pixel $\tilde{\mathbf{p}}$ be $\{\tilde{\mathbf{p}}_k\}_{k=1}^4$ with bilinear weights $\{\omega_k\}_{k=1}^4$. The interpolation can be written as,
\begin{equation}
    \label{eq::sn_sn}
    \mathbf{N}_\text{Direct}(\mathbf{p}):=\left<\mathbf{N}_s(\tilde{\mathbf{p}})\right>_{\tilde{\mathbf{p}}}=\sum_{k}\omega_k\mathbf{N}_s(\tilde{\mathbf{p}}_k)\propto\sum_{k}\omega_k(\mathbf{v}_k^x\times\mathbf{v}_k^y),
\end{equation}
where we denote $\mathbf{N}_s$ as surface normal from source context.
\par In depth interpolation-based method, depth is first reprojected to target view with modified backward warping and then interpolated,
\begin{equation}
    \tilde{\mathbf{D}}_s(\mathbf{p}):=\left<\tilde{\mathbf{D}}_s(\tilde{\mathbf{p}})\right>_{\tilde{\mathbf{p}}}=\sum_{k}\omega_k\tilde{\mathbf{D}}_s(\tilde{\mathbf{p}}_k),
    \label{eq::depth_inter}
\end{equation}
where $\tilde{\mathbf{D}}_s$ denotes depth from source context but viewed in target perspective.
The neighboring depths are similarly interpolated as,
\begin{equation}
    \tilde{\mathbf{D}}_s^x(\mathbf{p}^x)=\sum_{k}\omega_k\tilde{\mathbf{D}}_s(\tilde{\mathbf{p}}_k^y),~
    \tilde{\mathbf{D}}_s^y(\mathbf{p}^y)=\sum_{k}\omega_k\tilde{\mathbf{D}}_s(\tilde{\mathbf{p}}_k^x)
\end{equation}
\par The corresponding tangent vectors have now become,
\begin{equation}
    \tilde{\mathbf{v}}^x=\overrightarrow{\mathbf{P}\mathbf{P}^x}=\tilde{D}_s^x\mathbf{K}^{-1}\mathbf{p}^x-\tilde{D}_s\mathbf{K}^{-1}\mathbf{p}.
    \label{eq::tangent_vector_depth}
\end{equation}
where $\tilde{D}_s^x=\tilde{\mathbf{D}}_s^x(\mathbf{p}^x)$, $\tilde{D}_s^y=\tilde{\mathbf{D}}_s^x(\mathbf{p}^y)$, and $\tilde{\mathbf{v}}^x$ denotes offset vector calculated with reprojected depth of source context.
\par Substituting (\ref{eq::depth_inter}) into (\ref{eq::tangent_vector_depth}) yields,
\begin{equation}
    \tilde{\mathbf{v}}^x=\sum_{k}\omega_k(\tilde{D}_s^x\mathbf{K}^{-1}\mathbf{p}_k^x-\tilde{D}_s\mathbf{K}^{-1}\mathbf{p})=\sum_{k}\omega_k\tilde{\mathbf{v}}_k^x
    \label{eq::tangent_vector_depth_new}
\end{equation}
\par Re-substituting (\ref{eq::tangent_vector_depth_new}) into (\ref{eq::norm}), the resulting surface normal reprojected with interpolated-depth can be formulated as, 
\begin{equation}
    \begin{aligned}
        \mathbf{N}_\text{Depth}(\mathbf{p})&\propto \left(\sum_{k}\omega_k\tilde{\mathbf{v}}^x_k\right)\times\left(\sum_{k}\omega_k\tilde{\mathbf{v}}^y_k\right)\\
        &=\sum_{i,j}\omega_i\omega_j(\tilde{\mathbf{v}}_i^x\times\tilde{\mathbf{v}}_j^y),
    \end{aligned}
    \label{eq::sn_depth}
\end{equation}
wher $i$ and $j\in\{1,2,3,4\}$.
\par
\begin{figure}[h]
    \centering
    \includegraphics[width=1\linewidth]{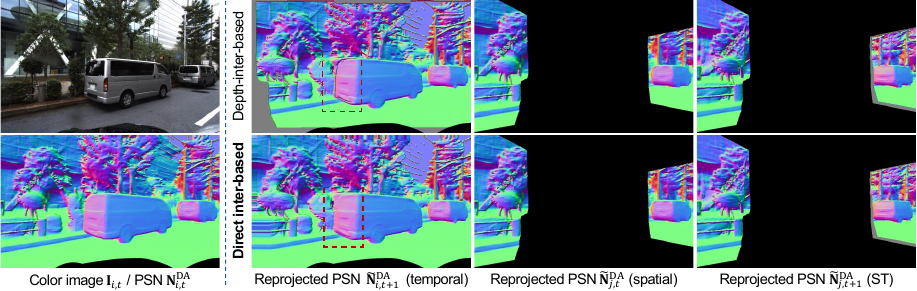}
    \caption{Comparison of depth and direct interpolation-based surface normal reprojection method. PSN implies pseudo surface normal. ST implies spatial-temporal context.}
    \label{fig::SN_warp}
\end{figure}

 Comparing (\ref{eq::sn_sn}) and (\ref{eq::sn_depth}), we observe that the direct interpolation-based reprojection method produces a weighted sum of surface normals from four neighboring pixels. In contrast, the depth interpolation-based method generates surface normals from combinations of offset vectors when $i \neq j$. These resulting normals are newly introduced and do not correspond to physically valid surface orientations. Consequently, $\mathbf{N}_\text{Direct}$ is generally smoother, whereas $\mathbf{N}_\text{Depth}$ tends to be significantly noisier and less reliable for supervision. This discrepancy is further amplified in regions with depth discontinuities, occlusion boundaries, and high curvature. Therefore, we do not adopt the latter as a supervisory signal in our framework. In Fig. \ref{fig::SN_warp}, we present an example of comparison between these two reprojection methods.

\subsection{Ground Plane-Aware Camera Height Regularization}
\par Camera height has been widely adopted as a scaling factor to enable metric monocular depth estimation in prior works like \cite{xue2020toward}\cite{wagstaff2021self}, where the ratio between the groundtruth and the estimated camera height is used to recover the absolute scale. However, these methods typically rely on the estimated depth to derive surface normals and subsequently detect the ground plane, which can be unstable during the early stages of training due to inaccurate depth estimation. 
\par In this work, we leverage DA to perform ground plane detection in a label-free manner. Concretely, we re-adopt the surface normal estimation approach in previous subsection and identify ground points whose surface normals are approximately aligned with the $y$-axis of the camera coordinate system, indicating consistency with the ground plane orientation. Specifically, we define $\mathbf{n}_y = [0,1,0]^\top$ and filter ground points based on cosine similarity as follows,
\begin{equation}
    s(\mathbf{p})=\angle(\mathbf{n}_y,\mathbf{N}^\text{DA}(\mathbf{p}))=\arccos\frac{\mathbf{n}_y\cdot \mathbf{N}^\text{DA}(\mathbf{p})}{\Vert\mathbf{n}_y\Vert\Vert\mathbf{N}^\text{DA}(\mathbf{p})\Vert}.
\end{equation}
\par To account for potential inaccuracies in surface normal estimation and deviations from perfect perpendicularity between the $y$-axis and the ground plane, a point $\mathbf{p}$ is classified as a ground point in the image plane if the corresponding cosine similarity lies within a predefined threshold $s_\text{thr}$ and and its 3D projection lies below the camera.
% and $y_\mathbf{P}$.Meanwhile, to avoid mistakenly filtering out points that lie on planes parallel to the ground but are located above the ground and below the camera, we further constrain $y_\mathbf{P}$ to be greater than $h_\text{GT}-h_\text{thr}$, where $h_\text{GT}$ is measurement of camera height, and $h_\text{thr}$ is a pre-defined height threshold. 
This condition can be formulated as follows,
\begin{equation}
    \boldsymbol{\mathcal{M}}_g^\text{DA}=\{\mathbf{p}||s(\mathbf{p})|<s_\text{thr},~(\mathbf{P})_y>0\}.
\end{equation}
\par Subsequently, we compute a pixel-wise camera height map using the estimated depth. Following the approach proposed in \cite{xue2020toward}, the camera height at each point can be estimated by projecting the vector $\overrightarrow{\mathbf{O}\mathbf{P}}$ onto the direction of the surface normal at point $\mathbf{p}$ as,
\begin{equation}
    \hat{\boldsymbol{h}}(\mathbf{p})=\hat{\mathbf{N}}(\mathbf{p})^\top\overrightarrow{\mathbf{O}\mathbf{P}}.
\end{equation}
\par With estimated camera heights and detected ground plane, the camera height regularization can thus be formulated as,
\begin{equation}
    \mathcal{L}_\text{CH}=\boldsymbol{\mathcal{M}}_g^\text{DA}\cdot\Vert \hat{\boldsymbol{h}}-h_\text{GT}\Vert_1,
\end{equation}
where $h_\text{GT}$ indicates measurement of camera height.
\par See Fig. \ref{fig::overview}(b) for visualization of ground-plane detection and camera height estimation examples.

\subsection{Pose Estimation and Cross-Vehicle Pose Consistency}
\par
\begin{figure}[h]
    \centering
    \includegraphics[width=1\linewidth]{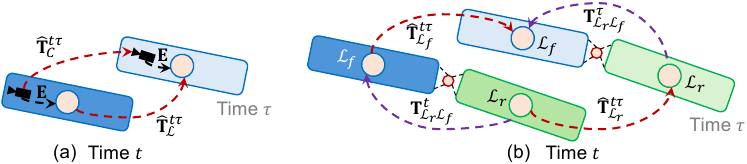}
    \caption{(a) Within- and (b) cross-vehicle pose consistency.}
    \label{fig::vehicle_consis}
\end{figure}

 Accurate pose estimation is essential for reliable pixel warping and subsequent view synthesis. In surround-view settings, prior studies constrain camera motion either through joint pose estimation or by introducing pose consistency regularization. In this work, we adopt the joint motion estimation strategy proposed in \cite{liugeosurdepth} for each vehicle and further enforce cross-vehicle consistency based on articulation characteristics. Specifically, we feed the surround-view images of each vehicle into the pose network to estimate the corresponding joint poses as,
\begin{equation}
    \hat{\mathbf{T}}_{\mathcal{L}_k}^{t\tau}=\mathcal{P}_\text{de}\left(\sum_{i\in\mathcal{C}_k}\omega_i\cdot\mathcal{P}_\text{en}(\mathbf{I}_i^t,\mathbf{I}_i^{\tau})\right),~k\in\{f,r\}
\end{equation}
where $\mathcal{P}_\text{en}$ and $\mathcal{P}_\text{de}$ denote pose encoder and decoder, respectively. $\mathcal{C}_k$ represents the set of cameras mounted on the front or rear vehicle. $\omega_i$ denotes the learnable parameter of the adaptive joint motion learning module proposed in \cite{liugeosurdepth}. The pose of each individual camera is derived by composing the estimated joint motion with its corresponding calibrated extrinsics through within-vehicle consistency (see Fig. \ref{fig::vehicle_consis}(a)). For example, the pose of camera $C_5$ on the front vehicle is computed as $\hat{\mathbf{T}}_{C_5}^{t\tau}=\mathbf{E}_{C_5}^{-1}\hat{\mathbf{T}}_{\mathcal{L}_f}^{t\tau}\mathbf{E}_{C_5}$.
\par Given the joint estimated motion of each vehicle, we define and formulate the cross-vehicle pose error and the corresponding consistency loss (see Fig. \ref{fig::vehicle_consis}(b)) as,
\begin{equation}
    \mathbf{T}_e=(\hat{\mathbf{T}}_{\mathcal{L}_f}^{t\tau}\mathbf{T}^t_{\mathcal{L}_r\mathcal{L}_f})^{-1}(\mathbf{T}^{\tau}_{\mathcal{L}_r\mathcal{L}_f}\hat{\mathbf{T}}_{\mathcal{L}_r}^{t\tau}),
\end{equation}
\begin{equation}
    \mathcal{L}_\text{VPC}=\lambda_R\Vert\mathbf{R}_e-\mathbf{I}\Vert_\text{Frob}+\lambda_t\Vert \mathbf{t}_e \Vert_2,
\end{equation}
where $\mathbf{R}_e$ and $\mathbf{t}_e$ denote the rotation and translation parts of $\mathbf{T}_e$, respectively, and $\lambda_R$ and $\lambda_t$ are weighting coefficients.

\subsection{Loss Function}
\par The overall loss function is composed of two parts: baseline losses and the newly proposed losses. The baseline losses include within-vehicle photometric loss $\mathcal{L}_p^\text{WV}$, smoothness loss $\mathcal{L}_s$, spatial depth consistency $\mathcal{L}_\text{SDC}$ proposed by \cite{cvcdepth}, and surface normal consistency $\mathcal{L}_\text{NC}$ and disparity smoothness consistency $\mathcal{L}_\text{DSC}$ proposed by \cite{liugeosurdepth}. The newly proposed losses comprise cross-view context enriched photometric loss $\mathcal{L}_p^\text{CV}$, cross-view spatial surface normal consistency $\mathcal{L}_\text{SNC}$, cross-view pseudo surface normal consistency $\mathcal{L}_\text{PNC}$, ground plane-aware camera height regularization $\mathcal{L}_\text{CH}$ and cross-vehicle pose consistency $\mathcal{L}_\text{VPC}$. Notably, the proposed cross-vehicle context enrichment is also employed in $\mathcal{L}_\text{SDC}$, $\mathcal{L}_\text{SNC}$ and $\mathcal{L}_\text{PNC}$. 
\section{Experiments}
\label{sec::experiment}
\subsection{Implementation Details}
\subsubsection{Experiment platform and dataset.} 
\begin{figure}[h]
    \centering
    \includegraphics[width=1\linewidth]{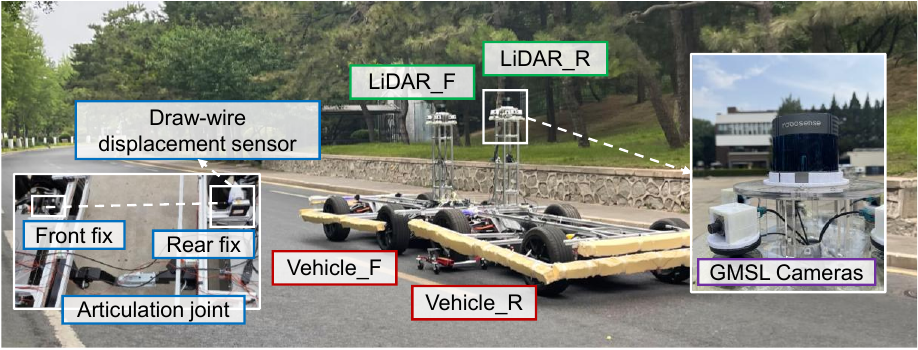}
    \caption{Our self-established experiment platform.}
    \label{fig::platform}
\end{figure}

As no public dataset is available for articulated vehicles, we built an experiment platform and collected a dataset over it. As illustrated in Fig. \ref{fig::abstract} and Fig. \ref{fig::platform}, we simulated a articulated vehicle with two independent chassis. Each chassis is equipped with 5 GMSL monocular cameras with FoV of $100^\circ$ and a 32-beam LiDAR.
% and an integrated IMU/RTK module for localization.
The two chassis are connected via a ball joint with negligible clearance. The articulation angle between two vehicles is measured with a mechanical sensor. All sensor data were recorded using ROS system and subsequently processed offline for multi-sensor time synchronization.
\par The dataset contains $4k$ data samples. Each sample includes 10 undistorted monocular images, resulting in a total of $40k$ images. It also provides 2 set of LiDAR pointclouds, which are used to generate 10 views of projected depth groundtruth, along with one articulation angle measurement. The dataset is split into $3.2k$ samples for training and $0.8k$ samples for validation and evaluation. 
\subsubsection{Training.} 
\begin{figure}[t]
    \centering
    \includegraphics[width=1\linewidth]{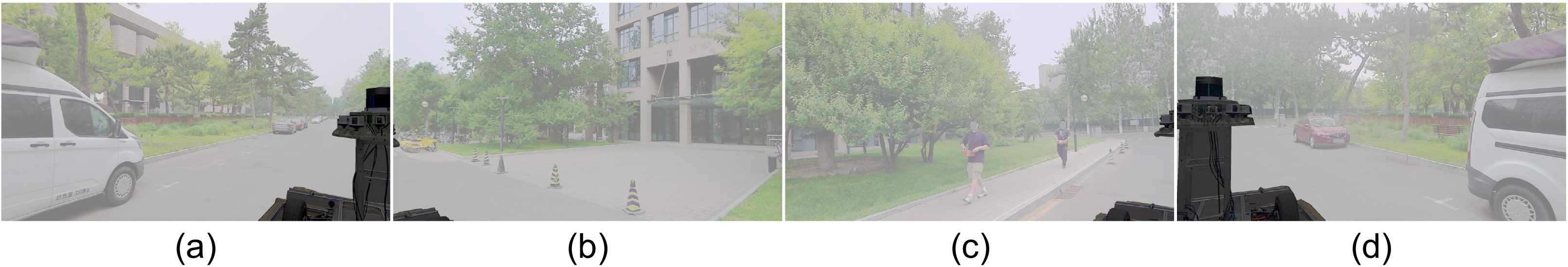}
    \caption{Self-occlusion masks overlaid on images: (a) $C_2$, (b) $C_3$, (c) $C_7$, (d) $C_8$ are four cameras that have visible occlusion components on the other vehicle under certain motion cases.}
    \label{fig::self_occlusion}
\end{figure}

Our framework was implemented in PyTorch and trained on four NVIDIA RTX A6000 GPUs. We utilized depth and pose network proposed by \cite{liugeosurdepth}. Input images were downsampled to $(384, 640)$, and adjacent frames ($\tau\in {t-1, t+1}$) were used as temporal context during training. We optimized the model using the Adam optimizer with $\beta_1 = 0.9$ and $\beta_2 = 0.999$, a learning rate of $1 \times 10^{-4}$, for 20 epochs. The batch size was set to 1 per GPU. Self-occlusion masks and reprojection masks were applied to exclude invalid pixels from loss computation. Fig. \ref{fig::self_occlusion} presents examples of the self-occlusion annotation results.
\subsubsection{Evaluation.} We evaluate metric depth up to a maximum range of 100 m. For quantitative comparison, we adopted the depth evaluation metrics proposed in \cite{eigen2014depth}. No horizontal-flip post-processing \cite{godard2019digging} was applied during evaluation. 

\subsection{Experiment Results}
\subsubsection{Monocular depth estimation and passenger vehicle surround-view depth estimation.} We first conduct experiments on the KITTI \cite{kitti} dataset by adding $\mathcal{L}_\text{PNC}^\text{T}$ to loss function of GeoDepth \cite{liugeosurdepth}. Results in Table \ref{table::result} show a slight improvement in depth evaluation metrics. This limited gain is expected, as conventional SfM and geometric constraints already provide strong supervision and KITTI dataset itself is relatively less challenging. Nevertheless, the results still indicate the effectiveness of the proposed formulation.
\par Meanwhile, experiments were also conducted on surround-view depth estimation datasets DDAD \cite{ddad} and nuScenes \cite{nuscenes}, which are collected on passenger vehicles. Concretely, we modified GeoSurDepth by replacing $\tilde{\mathcal{L}}_\text{NC}$ with $\mathcal{L}_\text{SNC}$ proposed in this work and adding $\mathcal{L}_\text{PNC}$ to the overall loss function. Results listed in Table \ref{table::result} indicate main improvement in metrics like Abs Rel and Sq Rel. This observation is consistent with the nature of geometry-based constraints. Specifically, when the effective baseline, either through temporal or spatial transformation, is small, the induced disparity becomes limited, particularly for distant regions. As a consequence, far-range pixels exhibit minimal geometric variation, and even after rotation compensation, the induced changes in surface normals remain minimal, thereby limiting the impact on far-distance regions. In contrast, near-range regions benefit more from the introduced geometric constraints, leading to more noticeable improvements.
\subsubsection{Articulated vehicle surround-view depth estimation.} Subsequently, experiments were conducted on our self-collected dataset, comparing our method with recent SoTA baselines. To adapt these baselines to the articulated vehicle setup, several modifications were made. For CVCDepth \cite{cvcdepth}, we used features from side cameras $C_6$ (front vehicle) and $C_4$ (rear vehicle) to estimate the motion of each vehicle. Although the original work uses front-view cameras for pose estimation, using $C_5$ (front-view) and $C_0$ (rear-view) in our dataset led to unstable training and non-convergent depth estimations, likely due to differences in environmental contexts and camera sensitivity to motion cues across datasets. For GeoSurDepth \cite{liugeosurdepth}, we implemented adaptive joint motion learning for both vehicles. Notably, modified baselines did not leverage enriched cross-vehicle contexts during training.
\par
\begin{figure}[t]
    \centering
    \includegraphics[width=1\linewidth]{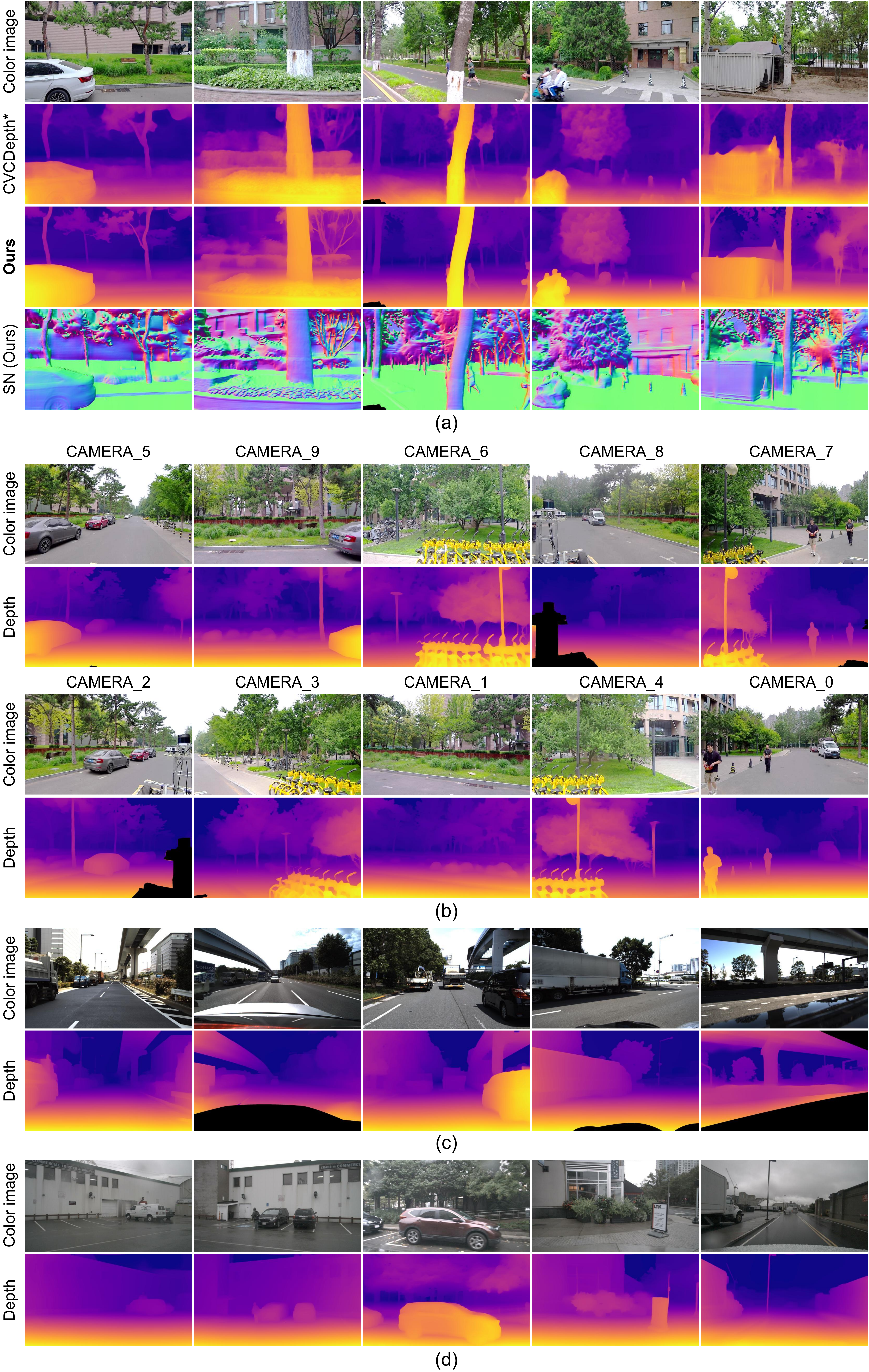}
    \caption{Visualization on \textit{val} split: (a) Examples of depth and surface normal estimation on self-collected dataset; (b) Example of surround-view depth estimation on test split of self-collected dataset; (c) Examples of direct inference on DDAD; (d) Examples of direct inference on nuScenes.}
    \label{fig::result}
\end{figure}

 Table \ref{table::result} shows superior performance of our method in comparison to other modified baselines. Qualitative results in Fig. \ref{fig::result}(a)(b) further show that the proposed approach produces smooth depth maps with enhanced edge and object awareness. In addition, cross-dataset evaluation indicates strong generalization ability, where ArticuSurDepth performs well under zero-shot inference on the DDAD and nuScenes datasets (Fig. \ref{fig::result}(c)(d)).
\begin{table}[t]
\setlength{\tabcolsep}{1pt}
\centering
\caption{Depth estimation on KITTI \cite{kitti}, DDAD \cite{ddad}, nuScenes \cite{nuscenes} and our own self-collected datasets (Results on KITTI are scale-ambiguous evaluation. $\dag$ indicates our modified GeoDepth or GeoSurDepth with proposed losses. \# implies reproduced results by VFDepth \cite{vfdepth}. * indicates modified framework for adaptation to articulated vehicle pattern. Results best in \textbf{bold}, second best \underline{underlined}).}
\label{table::result}
\scalebox{0.95}{
\begin{tabular}{c|*{7}{c}}
\toprule
Method & \cellcolor{red!8}{Abs Rel$\downarrow$} & \cellcolor{red!8}{Sq Rel$\downarrow$} & \cellcolor{red!8}{RMSE$\downarrow$} & \cellcolor{red!8}{RMSE$_\text{log}$$\downarrow$} & \cellcolor{cyan!8}{$\delta_1$$\uparrow$} & \cellcolor{cyan!8}{$\delta_2$$\uparrow$} & \cellcolor{cyan!8}{$\delta_3$$\uparrow$}\\
\midrule

% KITTI
\multicolumn{8}{c}{KITTI \cite{kitti} (192$\times$ 640)}\\
\midrule
MonoViT\cite{monovit} & 0.099 & 0.708 & 4.372 & 0.175 & 0.900 & 0.967 & 0.984\\
EDS-Depth\cite{yu2025eds} & \underline{0.095} & \textbf{0.619} & 4.184 & 0.170 & 0.905 & 0.969 & \underline{0.985}\\
GeoDepth\cite{liugeosurdepth} & \textbf{0.094} & 0.641 & \underline{4.175} & \underline{0.167} & \underline{0.906} & \underline{0.970} & \textbf{0.986}\\
\cellcolor{gray!15}{\textbf{Ours}}$\dag$ & \textbf{0.094} & \underline{0.628} & \textbf{4.134} & \textbf{0.165} & \textbf{0.908} & \textbf{0.971} & \textbf{0.986}\\
\midrule

% DDAD
\multicolumn{8}{c}{DDAD \cite{ddad} (384$\times$ 640)}\\
\midrule
FSM\#\cite{fsm} & 0.228 & 4.409 & 13.433 & 0.342 &  0.687 & 0.870 & 0.932 \\
VFDepth \cite{vfdepth} & 0.218 & 3.660 & 13.327 & 0.339 & 0.674 & 0.862 & 0.932 \\
SurroundDepth\cite{surrounddepth} & 0.208 & 3.371 & 12.977 & 0.330 & 0.693 & 0.871 & 0.934 \\
CVCDepth\cite{cvcdepth} & 0.210 & 3.458 & 12.876 & - & \underline{0.704} & - & - \\
GeoSurDepth\cite{liugeosurdepth} & \underline{0.176} & \underline{2.738} & \textbf{11.520} & \textbf{0.280} & \textbf{0.763} & \textbf{0.912} & \textbf{0.957} \\
\cellcolor{gray!15}{\textbf{Ours}}$\dag$ & \textbf{0.173} & \textbf{2.713} & \underline{11.726} & \underline{0.282} & \textbf{0.763} & \underline{0.910} & \underline{0.956}\\
\midrule

% nuScenes
\multicolumn{8}{c}{nuScenes \cite{nuscenes} (352$\times$ 640)}\\
\midrule
FSM\#\cite{fsm} & 0.319 & 7.534 & 7.860 & 0.362 & 0.716 & 0.874 & 0.931\\
VFDepth\cite{vfdepth} & 0.289 & 5.718 & 7.551 & 0.348 & 0.709 & 0.876 & \underline{0.932}\\
SurroundDepth\cite{surrounddepth} & 0.280 & \textbf{4.401} & 7.467 & 0.364 & 0.661 & 0.844 & 0.917 \\
CVCDepth\cite{cvcdepth} & 0.258 & \underline{4.540} & 7.030 & - & 0.756 & - & - \\
GeoSurDepth\cite{liugeosurdepth} & \underline{0.215} & 4.845 & \textbf{6.157} & \textbf{0.282} & \textbf{0.823} & \textbf{0.922} & \textbf{0.954} \\
\cellcolor{gray!15}{\textbf{Ours}}$\dag$ & \textbf{0.212} & 4.837 & \underline{6.288} & \underline{0.285} & \underline{0.822} & \underline{0.921} & \textbf{0.954}\\
\midrule

% Own
\multicolumn{8}{c}{Self-collected (384$\times$ 640)}\\
\midrule
CVCDepth*\cite{cvcdepth} & 0.234 & 1.878 & 7.164 & 0.326 & 0.587 & 0.864 & 0.948\\
GeoSurDepth*\cite{liugeosurdepth} & \underline{0.196} &	\underline{1.269} & \underline{6.278} & \underline{0.266}	& \underline{0.639} & \underline{0.922} & \underline{0.971} \\
\cellcolor{gray!15}{\textbf{Ours}} & \textbf{0.185} & \textbf{1.153} & \textbf{5.973} & \textbf{0.250} & \textbf{0.670} & \textbf{0.931} & \textbf{0.976}\\
\bottomrule
\end{tabular}
}
\end{table}

\subsection{Ablation Studies}
\par In this section, we report ablation studies of proposed methods on self-collected dataset in terms of metric depth estimation.
\subsubsection{Cross-vehicle context enrichment.} Table \ref{table::ablation_cv_context} shows that incorporating type-1\&2 cross-vehicle contexts for spatial cues improves model performance. This is quite intuitive, as these two types of spatial contexts provide broader overlapping areas. In contrast, type-0 offers relatively large overlapping FoV only when the vehicle exhibits significant articulation angles, such as during turning.
\begin{table}[t]
\setlength{\tabcolsep}{2pt}
\centering
\textbf{\caption{Ablation study on cross-vehicle context enrichment ($S_i$ indicate type-$i$ cross-vehicle spatial contexts).}
\label{table::ablation_cv_context}}
\scalebox{0.95}{
\begin{tabular}{ccc|*{7}{c}}
\toprule
$S_0$ & $S_1$ & $S_2$ & \cellcolor{red!8}{Abs Rel$\downarrow$} & \cellcolor{red!8}{Sq Rel$\downarrow$} & \cellcolor{red!8}{RMSE$\downarrow$} & \cellcolor{red!8}{RMSE$_\text{log}$$\downarrow$} & \cellcolor{cyan!8}{$\delta_1$$\uparrow$} & \cellcolor{cyan!8}{$\delta_2$$\uparrow$} & \cellcolor{cyan!8}{$\delta_3$$\uparrow$}\\
\midrule
\ding{55} & \ding{55} & \ding{55} & 0.192 & 1.235 & 6.237 & 0.259 & 0.656 & 0.921 & 0.973\\
\ding{51} & \ding{55} & \ding{55} & 0.190 & 1.193 & 6.069 & 0.255 & 0.661 & 0.926 & \underline{0.975}\\
\ding{55} & \ding{51} & \ding{55} & \underline{0.187} & \underline{1.165} & \underline{5.983} & \underline{0.251} & \underline{0.667} & \underline{0.929} & \textbf{0.976}\\
\ding{55} & \ding{55} & \ding{51} & 0.188 & 1.173 & 5.992 & 0.252 & 0.666 & \underline{0.929} & \underline{0.975}\\
\midrule
\ding{51} & \ding{51} & \ding{51} & \textbf{0.185} & \textbf{1.153} & \textbf{5.973} & \textbf{0.250} & \textbf{0.670} & \textbf{0.931} & \textbf{0.976}\\
\bottomrule
\end{tabular}
}
\end{table}

\subsubsection{Cross-view surface normal consistency.} Results in Table \ref{table::ablation_SNC} show that removing $\mathcal{L}_\text{PNC}$ leads to a degradation in depth estimation performance. As surround-view data of articulated vehicle provides richer multi-view geometric priors for structural learning, the proposed constraint can better exploit such information. This observation is further supported by experiments results on DDAD and nuScenes demonstrating consistent improvements when the baseline is enhanced with $\mathcal{L}_\text{PNC}$. Meanwhile, when replacing $\mathcal{L}_\text{SNC}$ with $\tilde{\mathcal{L}}_\text{NC}$ proposed in \cite{liugeosurdepth}, a slight performance drop is observed. We attribute this to the fact that our approach better preserves multi-view geometric self-consistency within the estimated geometry, thereby enforcing stronger geometric closure.
\begin{table}[t]
\setlength{\tabcolsep}{1.6pt}
\centering
\textbf{\caption{Ablation study on cross-view SN consistency (``D'' indicates depth interpolation-based SN reprojection.}
\label{table::ablation_SNC}}
\scalebox{0.95}{
\begin{tabular}{c|*{7}{c}}
\toprule
Method & \cellcolor{red!8}{Abs Rel$\downarrow$} & \cellcolor{red!8}{Sq Rel$\downarrow$} & \cellcolor{red!8}{RMSE$\downarrow$} & \cellcolor{red!8}{RMSE$_\text{log}$$\downarrow$} & \cellcolor{cyan!8}{$\delta_1$$\uparrow$} & \cellcolor{cyan!8}{$\delta_2$$\uparrow$} & \cellcolor{cyan!8}{$\delta_3$$\uparrow$}\\
\midrule
w/o $\mathcal{L}_\text{PNC}$ & 0.189 & 1.173 & 6.000 & 0.253 & 0.664 & 0.928 & \textbf{0.976}\\
with $\tilde{\mathcal{L}}_\text{NC}$ & \underline{0.186} & \underline{1.162} & \underline{5.981} & \textbf{0.250} & \underline{0.668} & \underline{0.930} & \textbf{0.976}\\
\textbf{Ours} & \textbf{0.185} & \textbf{1.153} & \textbf{5.973} & \textbf{0.250} & \textbf{0.670} & \textbf{0.931} & \textbf{0.976}\\
\bottomrule
\end{tabular}
}
\end{table}

\subsubsection{Ground-plane aware camera height regularization.} 
\begin{figure}[h]
    \centering
    \includegraphics[width=1\linewidth]{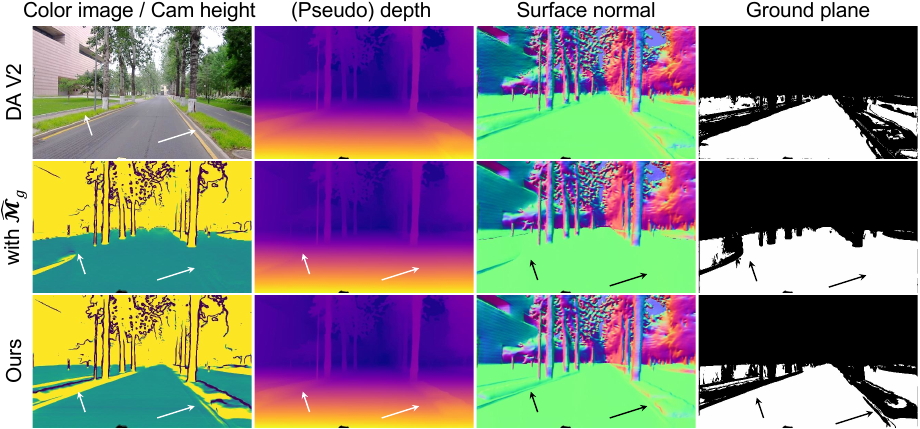}
    \caption{Example of ground plane detection or estimation, and camera height estimation (range of camera height estimation has been clipped for visualization).}
    \label{fig::cam_height_result}
\end{figure}

In this work, we utilize surface normals generated by DA to achieve stable and semantic-independent ground plane-aware camera height regularization. As shown in Table \ref{table::cam_height}, removing $\mathcal{L}_\text{CH}$ or replacing the ground plane mask with one generated from estimated surface normals leads to performance degradation. The visualization in Fig. \ref{fig::cam_height_result} further indicates that inaccurate surface normal estimation may result in false detection of ground points, which in turn regularizes the camera height using incorrect regions.
\begin{table}[h]
\setlength{\tabcolsep}{1.5pt}
\centering
\textbf{\caption{Ablation study on ground-plane aware camera height regularization ($\hat{\boldsymbol{\mathcal{M}}}_g$ indicates ground-plane detected with estimated depth and surface normal).}
\label{table::cam_height}}
\scalebox{0.95}{
\begin{tabular}{c|*{7}{c}}
\toprule
Method & \cellcolor{red!8}{Abs Rel$\downarrow$} & \cellcolor{red!8}{Sq Rel$\downarrow$} & \cellcolor{red!8}{RMSE$\downarrow$} & \cellcolor{red!8}{RMSE$_\text{log}$$\downarrow$} & \cellcolor{cyan!8}{$\delta_1$$\uparrow$} & \cellcolor{cyan!8}{$\delta_2$$\uparrow$} & \cellcolor{cyan!8}{$\delta_3$$\uparrow$}\\
\midrule
w/o $\mathcal{L}_\text{CH}$ & \underline{0.189} & \underline{1.254} & \underline{6.229} & \underline{0.258} & \underline{0.667} & \underline{0.921} & \underline{0.973}\\
with $\hat{\boldsymbol{\mathcal{M}}}_g$ & 0.202 & 1.454 & 6.491 & 0.268 & 0.640 & 0.911 & 0.969\\
\textbf{Ours} & \textbf{0.185} & \textbf{1.153} & \textbf{5.973} & \textbf{0.250} & \textbf{0.670} & \textbf{0.931} & \textbf{0.976}\\
\bottomrule
\end{tabular}
}
\end{table}

\subsubsection{Cross-vehicle pose consistency.} Table \ref{table::ablation_VPC} compares the effectiveness of $\mathcal{L}_\text{VPC}$ under two motion estimation schemes. The first performs joint pose estimation for front and rear LiDAR (vehicle), denoted as $J_f-J_r$. The second estimates the camera poses using features from $C_6$ and $C_4$ and then transforms them to front and rear LiDAR poses via extrinsics. The results show a slight drop in depth estimation performance when $\mathcal{L}_\text{VPC}$ is removed from the overall loss, indicating that bridging the two vehicles via motion in the temporal domain provides additional cross-vehicle geometric,  dynamic and spatial constraints that benefit depth estimation.
\begin{table}[h]
\setlength{\tabcolsep}{1.5pt}
\centering
\textbf{\caption{Ablation study on cross-vehicle pose consistency.}
\label{table::ablation_VPC}}
\scalebox{0.94}{
\begin{tabular}{c|c|c|c|c}
\toprule
\multirow{2}{*}{\vspace{-0.5em}Method} &\multicolumn{2}{c}{\cellcolor{red!8}{Abs Rel$\downarrow$}}\vline & \multicolumn{2}{c}{\cellcolor{cyan!8}{$\delta_1$$\uparrow$}} \\
\cmidrule{2-5}
& $J_f-J_r$ & $C_6-C_4$ & $J_f-J_r$ & $C_6-C_4$ \\
\midrule
\textbf{w} / w.o $\mathcal{L}_\text{VPC}$ & \textbf{0.185} / 0.188 & \textbf{0.187} / 0.189 & \textbf{0.670} / 0.665 & \textbf{0.665} / 0.664\\
\bottomrule
\end{tabular}
}
\end{table}

\section{Conclusion}
\label{sec::conclusion}
In this work, we presented ArticuSurDepth, a unified framework and experimental platform for self-supervised surround-view depth estimation in articulated vehicle systems. By exploiting cross-view, particularly cross-vehicle, geometric consistency as auxiliary supervisory cues, the proposed framework enhances depth learning under complex articulated configurations. The framework integrates spatial context enrichment across vehicles, geometric consistency constraints for 3D structural coherence, and motion coupling within the articulated system to improve depth estimation. Meanwhile, vision foundation model priors are incorporated to provide pseudo geometric regularization and facilitate metric depth learning without requiring additional labels. Extensive experiments on self-collected and public datasets demonstrate that ArticuSurDepth achieves competitive performance, demonstrating the effectiveness of modeling geometric consistency, especially cross-vehicle ones, for more accurate self-supervised surround-view depth estimation.

\bibliographystyle{IEEEtran}
\bibliography{reference.bib}

@inproceedings{cvcdepth,
  title={Towards Cross-View-Consistent Self-Supervised Surround Depth Estimation},
  author={Ding, Laiyan and Jiang, Hualie and Li, Jie and Chen, Yongquan and Huang, Rui},
  booktitle={2024 IEEE/RSJ International Conference on Intelligent Robots and Systems (IROS)},
  pages={10043--10050},
  year={2024},
  organization={IEEE}
}

@article{vfdepth,
  title={Self-supervised surround-view depth estimation with volumetric feature fusion},
  author={Kim, Jung-Hee and Hur, Junhwa and Nguyen, Tien Phuoc and Jeong, Seong-Gyun},
  journal={Advances in Neural Information Processing Systems},
  volume={35},
  pages={4032--4045},
  year={2022}
}

@article{fsm,
  title={Full surround monodepth from multiple cameras},
  author={Guizilini, Vitor and Vasiljevic, Igor and Ambrus, Rares and Shakhnarovich, Greg and Gaidon, Adrien},
  journal={IEEE Robotics and Automation Letters},
  volume={7},
  number={2},
  pages={5397--5404},
  year={2022},
  publisher={IEEE}
}

@inproceedings{surrounddepth,
  title={Surrounddepth: Entangling surrounding views for self-supervised multi-camera depth estimation},
  author={Wei, Yi and Zhao, Linqing and Zheng, Wenzhao and Zhu, Zheng and Rao, Yongming and Huang, Guan and Lu, Jiwen and Zhou, Jie},
  booktitle={Conference on robot learning},
  pages={539--549},
  year={2023},
  organization={PMLR}
}

@article{mcdp,
  title={Self-Supervised Multi-Camera Collaborative Depth Prediction With Latent Diffusion Models},
  author={Xu, Jialei and Liu, Xianming and Bai, Yuanchao and Jiang, Junjun and Ji, Xiangyang},
  journal={IEEE Transactions on Intelligent Transportation Systems},
  year={2025},
  publisher={IEEE}
}

@inproceedings{ddad,
  title={3d packing for self-supervised monocular depth estimation},
  author={Guizilini, Vitor and Ambrus, Rares and Pillai, Sudeep and Raventos, Allan and Gaidon, Adrien},
  booktitle={Proceedings of the IEEE/CVF conference on computer vision and pattern recognition},
  pages={2485--2494},
  year={2020}
}

@inproceedings{nuscenes,
  title={nuscenes: A multimodal dataset for autonomous driving},
  author={Caesar, Holger and Bankiti, Varun and Lang, Alex H and Vora, Sourabh and Liong, Venice Erin and Xu, Qiang and Krishnan, Anush and Pan, Yu and Baldan, Giancarlo and Beijbom, Oscar},
  booktitle={Proceedings of the IEEE/CVF conference on computer vision and pattern recognition},
  pages={11621--11631},
  year={2020}
}

@inproceedings{godard2019digging,
  title={Digging into self-supervised monocular depth estimation},
  author={Godard, Cl{\'e}ment and Mac Aodha, Oisin and Firman, Michael and Brostow, Gabriel J},
  booktitle={Proceedings of the IEEE/CVF international conference on computer vision},
  pages={3828--3838},
  year={2019}
}

@article{depthanythingv2,
  title={Depth anything v2},
  author={Yang, Lihe and Kang, Bingyi and Huang, Zilong and Zhao, Zhen and Xu, Xiaogang and Feng, Jiashi and Zhao, Hengshuang},
  journal={Advances in Neural Information Processing Systems},
  volume={37},
  pages={21875--21911},
  year={2024}
}

@inproceedings{depthanythingv1,
  title={Depth anything: Unleashing the power of large-scale unlabeled data},
  author={Yang, Lihe and Kang, Bingyi and Huang, Zilong and Xu, Xiaogang and Feng, Jiashi and Zhao, Hengshuang},
  booktitle={Proceedings of the IEEE/CVF conference on computer vision and pattern recognition},
  pages={10371--10381},
  year={2024}
}

@inproceedings{monovit,
  title={Monovit: Self-supervised monocular depth estimation with a vision transformer},
  author={Zhao, Chaoqiang and Zhang, Youmin and Poggi, Matteo and Tosi, Fabio and Guo, Xianda and Zhu, Zheng and Huang, Guan and Tang, Yang and Mattoccia, Stefano},
  booktitle={2022 international conference on 3D vision (3DV)},
  pages={668--678},
  year={2022},
  organization={IEEE}
}

@inproceedings{godard2017unsupervised,
  title={Unsupervised monocular depth estimation with left-right consistency},
  author={Godard, Cl{\'e}ment and Mac Aodha, Oisin and Brostow, Gabriel J},
  booktitle={Proceedings of the IEEE conference on computer vision and pattern recognition},
  pages={270--279},
  year={2017}
}

@article{eigen2014depth,
  title={Depth map prediction from a single image using a multi-scale deep network},
  author={Eigen, David and Puhrsch, Christian and Fergus, Rob},
  journal={Advances in neural information processing systems},
  volume={27},
  year={2014}
}

@inproceedings{zhou2017unsupervised,
  title={Unsupervised learning of depth and ego-motion from video},
  author={Zhou, Tinghui and Brown, Matthew and Snavely, Noah and Lowe, David G},
  booktitle={Proceedings of the IEEE conference on computer vision and pattern recognition},
  pages={1851--1858},
  year={2017}
}

@inproceedings{yang2021self,
  title={Self-supervised learning of depth inference for multi-view stereo},
  author={Yang, Jiayu and Alvarez, Jose M and Liu, Miaomiao},
  booktitle={Proceedings of the IEEE/CVF Conference on Computer Vision and Pattern Recognition},
  pages={7526--7534},
  year={2021}
}

@article{poggi2021synergies,
  title={On the synergies between machine learning and binocular stereo for depth estimation from images: A survey},
  author={Poggi, Matteo and Tosi, Fabio and Batsos, Konstantinos and Mordohai, Philippos and Mattoccia, Stefano},
  journal={IEEE Transactions on Pattern Analysis and Machine Intelligence},
  volume={44},
  number={9},
  pages={5314--5334},
  year={2021},
  publisher={IEEE}
}

@inproceedings{xue2020toward,
  title={Toward hierarchical self-supervised monocular absolute depth estimation for autonomous driving applications},
  author={Xue, Feng and Zhuo, Guirong and Huang, Ziyuan and Fu, Wufei and Wu, Zhuoyue and Ang, Marcelo H},
  booktitle={2020 IEEE/RSJ International Conference on Intelligent Robots and Systems (IROS)},
  pages={2330--2337},
  year={2020},
  organization={IEEE}
}

@article{yu2025eds,
  title={EDS-depth: Enhancing self-supervised monocular depth estimation in dynamic scenes},
  author={Yu, Shangshu and Wu, Meiqing and Lam, Siew-Kei and Wang, Changshuo and Wang, Ruiping},
  journal={IEEE Transactions on Intelligent Transportation Systems},
  year={2025},
  publisher={IEEE}
}

@article{liugeosurdepth,
  title={GeoSurDepth: Spatial Geometry-Consistent Self-Supervised Depth Estimation for Surround-View Cameras},
  author={Liu, Weimin and Wang, Wenjun and Meng, Joshua H},
  journal={arXiv preprint arXiv:2601.05839},
  year={2026}
}

@article{liu2025articubevseg,
  title={ArticuBEVSeg: Road Semantic Understanding and its Application in Bird's Eye View From Panoramic Vision System of Long Combination Vehicles},
  author={Liu, Weimin and Wang, Wenjun},
  journal={IEEE Robotics and Automation Letters},
  year={2025},
  publisher={IEEE}
}

@article{liu2026weak,
  title={Weak-Supervised Simultaneous Panoramic View Generation and Articulation Angle Estimation for Long Combination Vehicles},
  author={Liu, Weimin and Liu, Siyu and Wang, Wenjun and Meng, Joshua H and Sun, Zhaocong},
  journal={IEEE Transactions on Intelligent Transportation Systems},
  year={2026},
  publisher={IEEE}
}

@techreport{feng2019calibration,
  title={Calibration and stitching methods of around view monitor system of articulated multi-carriage road vehicle for intelligent transportation},
  author={Feng, Xiexing and Wei, LI and Wei, Tianyuan and Zhang, Yinglin and Cao, Libo},
  year={2019},
  institution={SAE Technical Paper}
}

@inproceedings{wagstaff2021self,
  title={Self-supervised scale recovery for monocular depth and egomotion estimation},
  author={Wagstaff, Brandon and Kelly, Jonathan},
  booktitle={2021 IEEE/RSJ International Conference on Intelligent Robots and Systems (IROS)},
  pages={2620--2627},
  year={2021},
  organization={IEEE}
}

@article{kitti,
  title={Vision meets robotics: The kitti dataset},
  author={Geiger, Andreas and Lenz, Philip and Stiller, Christoph and Urtasun, Raquel},
  journal={The international journal of robotics research},
  volume={32},
  number={11},
  pages={1231--1237},
  year={2013},
  publisher={Sage Publications Sage UK: London, England}
}

\end{document}